\pdfoutput=1

\documentclass[11pt]{article}

\usepackage{acl}

\usepackage{times}
\usepackage{latexsym}
\usepackage{inconsolata} 
\usepackage[T1]{fontenc}

\usepackage[utf8]{inputenc}

\usepackage{microtype}

\usepackage{bm}
\usepackage{booktabs}
\usepackage{siunitx}
\usepackage{graphicx}

\usepackage{enumitem}

\usepackage{amsmath}

\newcommand{\rpm}{\raisebox{.2ex}{$\scriptstyle\pm$}}

\title{On Isotropy Calibration of Transformers}

\author{Yue Ding$^2$, Karolis Martinkus$^1$, Dami\'an Pascual$^1$, Simon Clematide$^2$, Roger Wattenhofer$^1$\Thanks{  First three authors in alphabetic order} \\
   $^1$ETH Z\"{u}rich~\;~$^2$University of Z\"{u}rich\\
   \href{mailto:yue.ding@uzh.ch}{\texttt{yue.ding@uzh.ch}}~\;~ 
   \href{mailto:damianp@ethz.ch}{\texttt{damianp@ethz.ch}}~\;~ \href{mailto:martinkus@ethz.ch}{\texttt{martinkus@ethz.ch}}\\ \href{mailto:siclemat@cl.uzh.ch}{\texttt{siclemat@cl.uzh.ch}}~\;~
   \href{mailto:wattenhofer@ethz.ch}{\texttt{wattenhofer@ethz.ch}}
 }

\begin{document}
\maketitle
\begin{abstract}
Different studies of the embedding space of transformer models suggest that the distribution of contextual representations is highly anisotropic --- the embeddings are distributed in a narrow cone. Meanwhile, static word representations (e.g., Word2Vec or GloVe) have been shown to benefit from isotropic spaces. Therefore, previous work has developed methods to calibrate the embedding space of transformers in order to ensure isotropy. However, a recent study~\cite{cai2021isotropy} shows that the embedding space of transformers is locally isotropic, which suggests that these models are already capable of exploiting the expressive capacity of their embedding space. In this work, we conduct an empirical evaluation of state-of-the-art methods for isotropy calibration on transformers and find that they do not provide consistent improvements across models and tasks. These results support the thesis that, given the local isotropy, transformers do not benefit from additional isotropy calibration.
\end{abstract}

\section{Introduction}

The impressive performance of transformer models~\cite{vaswani2017attention} across almost all areas of Natural Language Processing (NLP) has sparked in-depth investigations of these models. 
A remarkable finding is that the contextual representations computed by transformers are strongly anistropic~\cite{contextual}, i.e., they are unevenly distributed and localized in a narrow cone of the embedding space. This discovery, labeled as the \emph{representation degeneration problem} by \citet{cosreg} is surprising since it suggests that most of the expressive capacity of these high-dimensional spaces is neglected by transformers.

Furthermore, previous work on static word representations, e.g., GloVE~\cite{pennington2014glove} or Word2Vec~\cite{word2vec}, established that isotropy is a desirable property in non-contextual embedding spaces~\cite{mu2017all}. Indeed, \citet{mu2017all} and \citet{liu2019unsupervised} showed that post-processing static word embeddings in order to increase isotropy improves their performance in downstream tasks.
Based on these results, recent work has developed methods to correct the anisotropy of the contextual representations generated by transformers~\cite{cosreg, spectrumcontrol, flowmodel}. These 
isotropy calibration methods have been reported to produce small gains in performance on some NLP tasks. 

However, in a recent study, \citet{cai2021isotropy} show that the space of contextual embeddings of transformers is locally isotropic. By analyzing low dimensional sub-spaces the authors identify isolated clusters and manifolds and argue that isotropy does exist in these manifolds. In the same line, \citet{luo2021catch} and \citet{kovaleva2021bert} find that in BERT~\cite{devlin2019bert} almost all of the embeddings present large values in the same two components of the embedding vector. These large components distort our understanding of the embedding spaces by making all the representations have high cosine similarity. 
In this work, we perform an extensive empirical evaluation of isotropy calibration methods across different tasks and models to determine if they provide consistent improvements. Our results question the utility of isotropy calibration in transformers, implicitly supporting the argument that transformers do already benefit from local isotropy~\cite{cai2021isotropy}.

\section{Related Work}

Since the appearance of the transformer architecture and its multiple variants, of which BERT~\cite{devlin2019bert} stands out as the most researched model, a lot of effort has been devoted to understanding their inner workings~\cite{rogers2020primer}. 
Unlike static word embeddings such as GloVE or Word2Vec, transformers build contextual embeddings, i.e., dynamic representations that aggregate information from other context words. These representations have sparked a lot of research interest. 
\citet{wu2020similarity} showed that different transformer architectures produce similar contextual representations. 
\citet{chronis2020bishop} studied the similarity and relatedness of contextual representations in the embedding spaces of BERT, 
while \citet{brunner2019identifiability} studied how identifiable the intermediate representations of BERT are with respect to the input. 
\citet{zhao2020quantifying} quantified the contextual knowledge of BERT and \citet{zhao2021non} analyzed the embedding spaces of BERT in order to quantify the non-linearity of its layers.


Following the discovery of anisotropy in transformers~\cite{cosreg, contextual}, different isotropy calibration methods have been developed to correct this phenomenon.
\citet{cosreg} and \citet{zhang2020revisiting} introduced regularization objectives that affect the embedding distances. \citet{isobn} presented a module inspired by batch-norm that regularizes the embeddings towards isotropic representations. \citet{spectrumcontrol} proposed to control the singular value decay of the output layer of transformers and \citet{flowmodel} used normalizing flows to map transformer embeddings to an isotropic space.
However, \citet{cai2021isotropy} show that contextual representations are locally isotropic and suggest that this property allows transformers to exploit their full expressive capacity, questioning the utility of isotropy calibration.

\begin{table*}[ht]
\centering
\resizebox{\textwidth}{!}{\begin{tabular}{@{}l*{11}{S[table-format=-3.4]}@{}}
\toprule
 & \multicolumn{1}{c}{SST-2} & \multicolumn{1}{c}{MRPC} & \multicolumn{1}{c}{CoLA} & \multicolumn{1}{c}{RTE} & \multicolumn{1}{c}{WNLI} & \multicolumn{1}{c}{STS-B} & \multicolumn{1}{c}{QNLI} & \multicolumn{2}{c}{MNLI} & \multicolumn{1}{c}{QQP}\\
\cmidrule(lr){2-2} \cmidrule(lr){3-3} \cmidrule(lr){4-4} \cmidrule(lr){5-5} \cmidrule(lr){6-6} \cmidrule(lr){7-7} \cmidrule(lr){8-8} \cmidrule(lr){9-10} \cmidrule(lr){11-11} 
{Model} & {Accuracy} & {F1} & {Mat. corr.} & {Accuracy} & {Accuracy} & {Pearson corr.} & {Accuracy} & {Match acc.} & {Mismatch acc.} & {Accuracy} \\
\midrule  

{BERT} & \makebox{\textbf{91.44 \rpm 0.52}} & \makebox{\textbf{88.80\rpm 0.99}} & \makebox{\textbf{53.16\rpm 1.82}} & \makebox{\textbf{58.97\rpm 1.82}} & \makebox{53.52\rpm 4.88} & \makebox{\textbf{80.86 \rpm 2.11}} & \makebox{88.78\rpm 0.57} & \makebox{81.02\rpm 0.17}&\makebox{81.78\rpm 0.40} & \makebox{89.31\rpm 0.06}\\
\emph{+Cosreg} & \makebox{90.71 \rpm 1.00} & \makebox{88.17 \rpm 0.38} & \makebox{46.94 \rpm 4.29} & \makebox{56.43 \rpm 5.16} & \makebox{50.23 \rpm 4.95} & \makebox{78.23 \rpm 2.19} & \makebox{\textbf{89.58 \rpm 0.19}} & \makebox{\textbf{81.20 \rpm 0.41}}&\makebox{\textbf{82.04 \rpm 0.21}} & \makebox{89.26 \rpm 0.10}\\
\emph{+Spectrum-Pol} & \makebox{90.86 \rpm 1.35} & \makebox{81.22 \rpm 0} & \makebox{0} & \makebox{49.58 \rpm 3.62} & \makebox{\textbf{56.34 \rpm 0}} & \makebox{NaN} & \makebox{81.24 \rpm 4.45} & \makebox{64.33 \rpm 27.80}&\makebox{64.76 \rpm 27.48} & \makebox{87.15 \rpm 2.23}\\
\emph{+Spectrum-Exp} & \makebox{91.21 \rpm 0.37} & \makebox{81.22 \rpm 0} & \makebox{0} & \makebox{50.90 \rpm 3.45} & \makebox{\textbf{56.34 \rpm 0}} & \makebox{NaN} & \makebox{86.42 \rpm 0.42} & \makebox{62.43 \rpm 24.97}&\makebox{63.12 \rpm 25.20} & \makebox{89.16 \rpm 0.45}\\
\emph{+Flow} & \makebox{91.09 \rpm 0.54} & \makebox{86.99 \rpm 0.89} & \makebox{51.19 \rpm 1.81} & \makebox{54.27 \rpm 1.46} & \makebox{48.36 \rpm 5.86} & \makebox{78.88 \rpm 3.46} & \makebox{86.21 \rpm 3.38} & \makebox{80.65 \rpm 0.46}&\makebox{ 81.15 \rpm 0.21} & \makebox{\textbf{89.36 \rpm 0.10}}\\
\midrule 

\addlinespace
{RoBERTa} & \makebox{\textbf{92.97 \rpm 0.63}} & \makebox{85.35 \rpm 8.52} & \makebox{\textbf{53.67 \rpm 3.32}} & \makebox{53.19 \rpm 0.55}  & \makebox{54.46 \rpm 0.81} & \makebox{\textbf{83.10 \rpm 2.87}} & \makebox{\textbf{91.00 \rpm 0.46}} & \makebox{85.16 \rpm 0.28}&\makebox{85.19 \rpm 0.15} & \makebox{89.85 \rpm 0.13}\\
\emph{+Cosreg} & \makebox{92.66 \rpm 0.23} & \makebox{\textbf{89.17 \rpm 2.28}} & \makebox{48.99 \rpm 5.61} & \makebox{\textbf{53.67 \rpm 1.16}} & \makebox{53.52 \rpm 1.41} & \makebox{28.44 \rpm 44.84} & \makebox{90.89 \rpm 0.19} & \makebox{\textbf{85.41 \rpm 0.09}}&\makebox{\textbf{85.64 \rpm 0.22} *} & \makebox{\textbf{89.87 \rpm 0.12}}\\
\emph{+Spectrum-Pol} & \makebox{88.08 \rpm 0.99} & \makebox{81.22 \rpm 0} & \makebox{0} & \makebox{52.71 \rpm 0} & \makebox{\textbf{57.28 \rpm 1.62} *} & \makebox{NaN} & \makebox{83.89 \rpm 2.46} & \makebox{50.63 \rpm 29.72}&\makebox{51.14 \rpm 29.29} & \makebox{81.76 \rpm 12.76}\\
\emph{+Spectrum-Exp} & \makebox{90.71 \rpm 1.09} & \makebox{81.22 \rpm 0} & \makebox{0} & \makebox{52.95 \rpm 0.42} & \makebox{56.34 \rpm 0} & \makebox{NaN} & \makebox{82.25 \rpm 3.14} & \makebox{84.46 \rpm 0.51}&\makebox{84.77 0.41} & \makebox{80.95 \rpm 13.89}\\ 
\midrule 

\addlinespace
{DistilBERT} & \makebox{88.23 \rpm 1.79} & \makebox{\textbf{87.97 \rpm 1.02}} & \makebox{44.11 \rpm 2.09} & \makebox{56.68 \rpm 0.62} & \makebox{51.17 \rpm 5.69} & \makebox{\textbf{23.63 \rpm 41.08}} & \makebox{\textbf{87.53 \rpm 0.13}} & \makebox{\textbf{78.84 \rpm 0.27}} & \makebox{\textbf{79.50 \rpm 0.32}} & \makebox{88.28 \rpm 0.25}\\ 
\emph{+Cosreg} & \makebox{88.53 \rpm 1.55} & \makebox{87.88 \rpm 1.36} & \makebox{\textbf{43.13 \rpm 0.85}} & \makebox{\textbf{58.24 \rpm 1.78}} & \makebox{52.11 \rpm 2.44} & \makebox{-0.50 \rpm 2.08} & \makebox{87.15 \rpm 0.84} & \makebox{78.69 \rpm 0.17} & \makebox{79.42 \rpm 0.28} & \makebox{88.38 \rpm 0.05}\\
\emph{+Spectrum-Pol} & \makebox{88.80 \rpm 0.37} & \makebox{81.22 \rpm 0} & \makebox{0} & \makebox{54.15 \rpm 2.50} & \makebox{\textbf{55.87 \rpm 0.81}} & \makebox{NaN} & \makebox{85.47 \rpm 0.96} & \makebox{78.39 \rpm 0.17} & \makebox{79.13 \rpm 0.05} & \makebox{\textbf{88.41 \rpm 0.43}}\\
\emph{+Spectrum-Exp} & \makebox{\textbf{88.92 \rpm 0.67}} & \makebox{81.22 \rpm 0} & \makebox{0} & \makebox{54.27 \rpm 2.71} & \makebox{\textbf{55.87 \rpm 0.81}} & \makebox{NaN} & \makebox{86.25 \rpm 0.80} & \makebox{78.38 \rpm 1.34} & \makebox{79.03 \rpm 0.34} & \makebox{88.12 \rpm 0.58}\\

\bottomrule  
\end{tabular}}
\caption{Performance for different models and calibration methods on GLUE; 
* denotes significantly better performance than the corresponding uncalibrated model ($p<0.05$, two-sample t-test). The NaN and 0 scores are caused by the model always predicting the same class.
}  
\label{tab:glue}
\end{table*}

\section{Isotropy Calibration Methods}
The output distribution of transformers is typically parameterized as a softmax function:
$$
P(\bm{Y}_i = \bm{y}_i | \bm{h}_i) = \frac{\exp(\bm{h}^T_i \bm{W}_{\mathcal{I}(\bm{y}_i)})}{\sum^N_{j=1}\exp(\bm{h}^T_i \bm{W}_{j})} \; ,
$$
where $\bm{W} \in \mathcal{R}^{N\times d}$ is the output weight matrix, $d$ is the embedding dimension, $N$ is the output size, 
$\bm{y}_i$ is the \textit{i}-th output, $\mathcal{I}(\bm{y}_i)$ is the index of $\bm{y}_i$ and $\bm{h}$ is the contextual embedding produced by the model. Since this constitutes a shared space between model embeddings $\bm{h} \in \bm{H}$ and output embeddings, 
isotropy at the output distribution can be enforced by calibrating either $\bm{H}$ or $\bm{W}$.

We experiment with three prominent methods for isotropy calibration on transformers:

\paragraph{Cosine Regularization.}\citet{cosreg} introduce a simple regularization term that minimizes the cosine similarity between any two output embeddings in order to increase the aperture of the cone that contains the embeddings. 
This regularization term is given by:

$$ \mathcal{R}_{cos}= \lambda_c \frac{1}{|\mathcal{V}|^2} \sum^n_{i} \sum^n_{j \neq i}\hat{\bm{w}}_i^T \hat{\bm{w}}_j  \; ,$$
\noindent where $\bm{w}_i$ is the embedding of the \textit{i}-th token in the vocabulary $\mathcal{V}$, $\hat{\bm{w}} = \frac{\bm{w}}{||\bm{w}||}$ and $\lambda_c$ is the regularization constant.

\paragraph{Spectrum Control.} 
\citet{spectrumcontrol} increase isotropy by mitigating the fast decay of the singular value distribution of the output matrix $\bm{W}$. They decompose $\bm{W}$ using Singular Value Decomposition (SVD), such that $\bm{W}  = \bm{U} \bm{\Sigma} \bm{V}^{T}$, where $\bm{\Sigma} \in \mathcal{R}^{d\times d}$ is the diagonal matrix of singular values. Then, they add a regularization term to guide the singular value distribution towards a pre-specified slow-decaying prior distribution. 
This term spreads the variance away from the first few dominating singular values, increasing the isotropy of the space.
They propose the following two regularization terms:
$$ \mathcal{R}_{pol}(\bm{\Sigma})= \lambda_p \sum^d_{k=1}(\sigma_k - c_1 k^\gamma)^2  \; , $$
for polynomial singular value decay; and $$ \mathcal{R}_{exp}(\bm{\Sigma})= \lambda_e \sum^d_{k=1}(\sigma_k - c_1 \exp (-c_2k^\gamma))^2  \; , $$
for exponential decay, where $\lambda_e$, $\lambda_p$, $c_1$ and $c_2$ are regularization constants, $\sigma_k$ is the \textit{k}-th largest singular value and $\gamma$ is a parameter which controls the rate of singular value decay.

\paragraph{Flow Model.}
\citet{flowmodel} propose a method that leverages normalizing flows to learn an invertible mapping $f_{\phi}^{-1}$ between the embedding space of the transformer model and an isotropic (Gaussian) space $\mathcal{Z}$.
First, an invertible flow model~\cite{glow} $f_{\phi}$ is trained to generate transformer embedding vectors $\bm{h}$ from Gaussian noise $\bm{z}$:
$$\bm{z} \sim p_\mathcal{Z}(\bm{z}),\; \bm{h} = f_{\phi}(\bm{z}) \; .$$  
Then, the model $f_{\phi}$ is inverted to map transformer embeddings $\bm{h}$ to the new (and isotropic) output embedding space $\mathcal{Z}$.

\section{Experiments}

We evaluate the impact of each of these calibration methods on state-of-the-art transformer models in three prominent areas of Natural Language Processing: language understanding, machine translation, and summarization. For all of the models, we use the implementation and fine-tuning parameters from HuggingFace~\cite{wolf2020transformers} (cf. Appendix~\ref{app:hyperparmeters}). We run each experiment three times and report the mean and standard deviation. Fine-tuning time is reported on a Nvidia Titan RTX GPU.

To characterize the isotropy of the output embedding space we adopt the $I_1$ and $I_2$ isotropy measures from \cite{spectrumcontrol}, with $I_1(\bm{W} ) \in [0,1]$ and $I_2(\bm{W} ) \geq 0$. 
Larger $I_1(\bm{W} )$ and smaller $I_2(\bm{W})$ indicate more isotropic embeddings (cf. App.~\ref{app:isomet} for details). %

\subsection{Language Understanding}

We consider three representative transformer models with different sizes, BERT-base~\cite{devlin2019bert}, 
RoBERTa~\cite{roberta}, 
and DistilBERT~\cite{sanh2020distilbert}. 
We evaluate these models on the development set of GLUE~\cite{glue}, a well-known benchmark for language understanding that consists of nine different tasks. 
Due to the high computational cost of flow calibration and the large number of tasks, we apply this method only on BERT to save resources.


In Table \ref{tab:glue} we report the performance per task of the calibrated and uncalibrated models. We observe the same pattern for all three models. In the overwhelming majority of cases, the calibrated models perform comparably to or worse than the uncalibrated ones, with calibration improving performance with statistical significance ($p<0.05$, two-sample t-test) only in RoBERTa for WNLI with exponential decay and MNLI mismatched with cosine regularization. More specifically, cosine regularization and flow calibration (in BERT) do not affect performance much, while spectrum control in some cases produces severe performance degradation or even prevents learning, e.g., CoLA and STS-B. 
Furthermore, flow calibration adds a large training overhead, requiring on average $4.2$ times more time per training epoch.

These results reveal that no isotropy calibration method yields consistently better performance than the uncalibrated models in language understanding tasks.

\begin{table*}[t]
\centering
\resizebox{\textwidth}{!}{\begin{tabular}{@{}l*{9}{S[table-format=-3.4]}@{}}
\toprule
 & \multicolumn{4}{c}{EN-RO} & \multicolumn{4}{c}{DE-EN}\\
\cmidrule(lr){2-5} \cmidrule(lr){6-9}
{Model} & {BLEU $(\uparrow)$} & {$I_1 (\uparrow)$} & {$I_2 (\downarrow)$} & {Time (min)} & {BLEU $(\uparrow)$} & {$I_1 (\uparrow)$} & {$I_2 (\downarrow)$} & {Time (min)}\\
\midrule  
{M-BART} & \makebox{\textbf{26.15 \rpm 0.08}} & \makebox{0.88 \rpm 0.01} & \makebox{0.60 \rpm 0} & \makebox{108 \rpm 0} & \makebox{22.81 \rpm 0.35} & \makebox{0.89 \rpm 0.01} & \makebox{0.60 \rpm 0} & \makebox{176 \rpm 0}\\  

\emph{+Cosreg} & \makebox{26.07 \rpm 0.10} & \makebox{0.88 \rpm 0.01} & \makebox{0.60 \rpm 0} & \makebox{110 \rpm 0} & \makebox{\textbf{23.03 \rpm 0.27}} & \makebox{0.89 \rpm 0.01} & \makebox{0.60 \rpm 0} & \makebox{188 \rpm 1} \\   

\emph{+Spectrum-Pol} & \makebox{22.94 \rpm 0.18} & \makebox{\textbf{1.00 \rpm 0}} & \makebox{\textbf{0.02 \rpm 0}} & \makebox{176 \rpm 2} & \makebox{16.27 \rpm 0.06} & \makebox{\textbf{1.00 \rpm 0}} & \makebox{\textbf{0.02 \rpm 0}} & \makebox{265 \rpm 0} \\      
\emph{+Spectrum-Exp} & \makebox{22.92 \rpm 0.05} & \makebox{\textbf{1.00 \rpm 0}} & \makebox{\textbf{0.02 \rpm 0}} & \makebox{170 \rpm 1} & \makebox{16.24 \rpm 0.12} & \makebox{\textbf{1.00 \rpm 0}} & \makebox{\textbf{0.02 \rpm 0}} & \makebox{230 \rpm 18} \\  
\midrule

M-BART (small dataset) & \makebox{\textbf{9.09 \rpm 1.02}} & \makebox{0.88 \rpm 0} & \makebox{\textbf{0.60 \rpm 0}} & \makebox{9 \rpm 0} & \makebox{\textbf{11.61 \rpm 2.25}} & \makebox{\textbf{0.88 \rpm 0}} & \makebox{\textbf{0.60 \rpm 0}} & \makebox{9 \rpm 0} \\  
\emph{+Flow} & \makebox{8.57 \rpm 2.52} & \makebox{\textbf{0.89 \rpm 0}} & \makebox{\textbf{0.60 \rpm 0}} & \makebox{95 \rpm 0} & \makebox{10.93 \rpm 0.70} & \makebox{\textbf{0.88 \rpm 0}} & \makebox{\textbf{0.60 \rpm 0}} & \makebox{96 \rpm 1} \\   
\bottomrule
\end{tabular}}
\caption{Multilingual BART performance, isotropy ($I_1$ and $I_2$) and fine-tuning time per epoch with different calibration methods for English - Romanian and German - English translation.
Due to computational cost, the flow method was tested only on a smaller version of the EN-RO dataset with $50\,000$ sentences.
}  
\label{tab:translation}
\end{table*}

\subsection{Machine Translation}

We test multilingual BART (M-BART) \cite{MBART} on English-Romanian and German-English WMT16 \cite{wmt} translation datasets. In Table \ref{tab:translation} we report BLUE scores, compute time, and the isotropy metrics, for the uncalibrated and calibrated models. 
To reduce the high computational cost of flow calibration, we apply this method only on a reduced version of $50\, 000$ samples for both tasks, English-Romanian and German-English translation.
As a reference, we also provide the scores of the uncalibrated model on the small datasets. 
We find, that while cosine regularization does not significantly affect either BLEU scores or isotropy metrics, both variants of spectrum control improve isotropy but produce a performance degradation of over $3$ and $5$ BLEU points in the English-Romanian and German-English tasks respectively, while requiring $25\%$ to $50\%$ more computation time. On the other hand, flow calibration yields comparable BLEU score to the uncalibrated model but requires on average $10.5$ times more computation per epoch. These results suggest a negative and counter-intuitive relation between isotropy and downstream performance: when isotropy increases, performance decreases. We observe a similar trend for language understanding in Appendix \ref{app:isoGLUE}.

Overall, and in line with the results in the previous section, isotropy calibration in machine translation tends to degrade performance and increase the computational budget.

\subsection{Summarization}

We evaluate BART \cite{bart} on the CNN/DM summarization task~\cite{CNNDailyMail}; again we use a reduced dataset ($20\,000$ articles) for flow calibration. The results in Table~\ref{tab:sum} 
show that none of the calibrated models performs significantly better than their uncalibrated counterparts in terms of ROUGE score~\cite{rouge} (cf. Appendix~\ref{app:sum}). Cosine regularization does not affect performance nor isotropy, while spectrum control improves isotropy ($I_1$ and $I_2$) at the cost of a small performance drop. The flow model performs comparably to uncalibrated BART but requires $5.5$ times more computation. 
Overall, we find no evidence that isotropy calibration provides gains in summarization.

\begin{table}[t]
\centering
\resizebox{\linewidth}{!}{\begin{tabular}{@{}l*{5}{S[table-format=-3.4]}@{}}
\toprule
 & \multicolumn{4}{c}{CNN / Daily Mail}\\
\cmidrule(lr){2-5}
{Model} & {R-1 $(\uparrow)$} & {$I_1 (\uparrow)$} & {$I_2 (\downarrow)$} & {Time (min)}\\
\midrule  
{BART} & \makebox{\textbf{38.21 \rpm 0.05}} & \makebox{0.95 \rpm 0.01} & \makebox{0.25 \rpm 0} & \makebox{246 \rpm 8}\\  

\emph{+Cosreg} & \makebox{\textbf{38.21 \rpm 0.05}} & \makebox{0.95 \rpm 0.01} & \makebox{0.25 \rpm 0} & \makebox{240 \rpm 8}\\ 

\emph{+Spectrum-Pol} & \makebox{37.36 \rpm 0.08} & \makebox{\textbf{0.99 \rpm 0}} & \makebox{\textbf{0.04 \rpm 0}} & \makebox{245 \rpm 20}\\      

\emph{+Spectrum-Exp} & \makebox{37.43 \rpm 0.08} & \makebox{\textbf{0.99 \rpm 0}} & \makebox{\textbf{0.04 \rpm 0}} & \makebox{230 \rpm 18}\\  

\midrule
{BART (small d.)} & \makebox{\textbf{36.56 \rpm 0.25}} & \makebox{\textbf{0.94 \rpm 0}} & \makebox{\textbf{0.25 \rpm 0}} & \makebox{17 \rpm 0 } \\  
\emph{+Flow} & \makebox{36.15 \rpm 0.30} & \makebox{\textbf{0.94 \rpm 0}} & \makebox{\textbf{0.25 \rpm 0}} & \makebox{95 \rpm 2} \\ 
\bottomrule
\end{tabular}}
\caption{ROUGE-1 score, isotropy ($I_1$ and $I_2$), and fine-tuning time per epoch with different calibration methods on BART for summarization. Due to computational cost, the flow calibration method was tested on a smaller version of the dataset.}
\label{tab:sum}
\end{table}

\section{Discussion}

Our extensive evaluation shows that none of the considered isotropy calibration methods produce consistent improvements over the uncalibrated models across tasks, domains and architectures. In fact, we observe a negative relation between isotropy calibration and downstream performance. The most aggressive method, i.e., spectrum control, produces the largest improvement in isotropy metrics as well as the most significant performance drop. On the other hand, the effect of cosine regularization and flow calibration is small in both, isotropy and performance.

According to \citet{cai2021isotropy}, the local isotropy of the embedding space of transformers may enable them to exploit their full expressive capacity. Furthermore, concurrent findings by \citet{luo2021catch} and \citet{kovaleva2021bert} reveal that certain components of the contextual embeddings consistently present very large magnitudes, which distort the cosine distances in the embedding space and questions their anisotropy.
This could explain why additional isotropy calibration does not consistently improve the performance of transformers in downstream tasks.

In light of our results, we discourage isotropy calibration of transformers as a means of improving downstream performance. However, we believe that further investigation of the embedding space of transformers may be beneficial to increase our ability to interpret these models and improve their architecture.

\bibliography{anthology,custom}
\bibliographystyle{acl_natbib}

\appendix
\onecolumn
\clearpage

\section{Isotropy Metrics}\label{app:isomet}

To characterize the isotropy of the output embedding space we adopt the $I_1$ and $I_2$ isotropy measures from \cite{spectrumcontrol}. 
$$
I_1(\bm{W})  = \frac{\min_{\bm{v} \in \bm{V}}Z(\bm{v})}{\max_{\bm{v} \in \bm{V}}Z(\bm{v})} \; ,
$$
is based on the observation by \cite{arora2016latent}, that the partition function $Z(\bm{v})=\sum_{i=1}^n \exp(\bm{v}^T \bm{w}_i)$ should be close to a constant for any unit vector $\bm{v}$ if the embedding matrix $\bm{W}$ is isotropic. Here, we abuse notation and $\bm{w_i} \in \bm{W}$ is the \textit{i}-th row of the embedding matrix $\bm{W}$. Following \cite{mu2017all} we use the set of eigenvectors of $\bm{W}^T\bm{W}$ as $\bm{V}$.
The second measure
$$
I_2(\bm{W})  = \sqrt{\frac{ \sum_{\bm{v} \in \bm{V}} (Z(\bm{v}) - \bar{Z}(\bm{v}))^2}{|V|\bar{Z}(\bm{v})^2}} \; ,
$$
is the sample standard deviation of the partition function $Z(\bm{v})$ normalized by its average $\bar{Z}(\bm{v})$.
This way, $I_1(\bm{W} ) \in [0,1]$ and $I_2(\bm{W} ) \geq 0$. 
Larger $I_1(\bm{W} )$ and smaller $I_2(\bm{W})$ indicate more isotropic embeddings. %

\section{Model Hyperparameter Configuration}\label{app:hyperparmeters}

For all the models used in his work we use the implementation from HuggingFace and follow their instructions for the hyperparameters. In particular, we use the following configurations:

\paragraph{BERT and DistilBERT.} Learning rate $2e^{-5}$ without scheduling, batch size $32$, $3$ training epochs for all GLUE tasks except for MRPC and WNLI, for which we train during $5$ epochs.

\paragraph{RoBERTa.} Learning rate of $1e^{-5}$ for all GLUE tasks except for SST-2 and STS-B, for which the learning rate is set to $1e^{-5}$, same number of epochs as for BERT and DistilBERT, batch size of $32$.

\paragraph{M-BART and BART.} Learning rate of $3e^{-5}$ with polynomial decay, batch size $48$, and $5$ training epochs.

\section{Isotropy Scores on GLUE}\label{app:isoGLUE}

Here, in Table \ref{tab:glue_isotropy}, we present the isotropy scores obtained in our evaluation of GLUE with BERT, RoBERTa, and DistilBERT, which were not included in the main text due to lack of space.

\begin{table*}[ht]
\centering
\resizebox{0.95\textwidth}{!}{\begin{tabular}{@{}l*{7}{S[table-format=-3.4]}@{}}
\toprule
 & \multicolumn{2}{c}{SST-2} & \multicolumn{2}{c}{MRPC} & \multicolumn{2}{c}{CoLA} 
 \\
\cmidrule(lr){2-3} \cmidrule(lr){4-5} \cmidrule(lr){6-7} 
{Model} & {$I_1 (\uparrow)$} & {$I_2 (\downarrow)$} & {$I_1 (\uparrow)$} & {$I_2 (\downarrow)$} & {$I_1 (\uparrow)$} & {$I_2 (\downarrow)$} 
\\
\midrule  
{BERT} & \makebox{0.91 \rpm 0.01} & \makebox{0.4 \rpm 0} & \makebox{0.91 \rpm 0.01} & \makebox{0.38 \rpm 0.01} & \makebox{0.91 \rpm 0.01} & \makebox{0.39 \rpm 0.01} 
\\  
\emph{+Cosreg} & \makebox{0.91 \rpm 0.2} & \makebox{0.39 \rpm 0.02} & \makebox{0.92 \rpm 0.01} & \makebox{0.39 \rpm 0.2} & \makebox{0.91 \rpm 0.01} & \makebox{0.39 \rpm 0.01} 
\\  
\emph{+Spectrum-Pol} & \makebox{\textbf{1.00 \rpm 0}} & \makebox{\textbf{0.007 \rpm 0.003}} & \makebox{\textbf{1.00 \rpm 0}} & \makebox{7$e^{-4}$ \rpm 3$e^{-4}$} & \makebox{\textbf{1.00 \rpm 0}} & \makebox{\textbf{6$e^{-4}$ \rpm 1$e^{-4}$}} 
\\
\emph{+Spectrum-Exp} & \makebox{0.99 \rpm 0.01} & \makebox{0.02 \rpm 0.02} & \makebox{\textbf{1.00 \rpm 0}} & \makebox{\textbf{6$e^{-4}$ \rpm 2$e^{-4}$}} & \makebox{\textbf{1.00 \rpm 0}} & \makebox{7$e^{-4}$ \rpm 3$e^{-4}$} 
\\
\emph{+Flow} & \makebox{0.92 \rpm 0.01} & \makebox{0.40 \rpm 0} & \makebox{0.91 \rpm 0.01} & \makebox{0.40 \rpm 0} & \makebox{0.91 \rpm 0.01} & \makebox{0.39 \rpm 0.01}
\\  
\midrule 
\addlinespace
{RoBERTa} & \makebox{0.91 \rpm 0.01} & \makebox{0.39 \rpm 0.01} & \makebox{0.92 \rpm 0.01} & \makebox{0.39 \rpm 0.01} & \makebox{0.91 \rpm 0.01} & \makebox{0.40 \rpm 0.01} 
\\  
\emph{+Cosreg} & \makebox{0.92 \rpm 0.01} & \makebox{0.40 \rpm 0.01} & \makebox{0.91 \rpm 0.01} & \makebox{0.39 \rpm 0.01} & \makebox{0.91 \rpm 0.01} & \makebox{0.40 \rpm 0.01} 
\\  
\emph{+Spectrum-Pol} & \makebox{\textbf{1.00 \rpm 0}} & \makebox{0.008 \rpm 0.002} & \makebox{\textbf{1.00 \rpm 0}} & \makebox{5$e^{-4}$ \rpm 4$e^{-4}$} & \makebox{\textbf{1.00 \rpm 0}} & \makebox{\textbf{5$e^{-4}$ \rpm 2$e^{-4}$}} 
\\  
\emph{+Spectrum-Exp} & \makebox{\textbf{1.00 \rpm 0}} & \makebox{\textbf{0.005 \rpm 0.004}} & \makebox{\textbf{1.00 \rpm 0}} & \makebox{\textbf{1$e^{-4}$ \rpm 2$e^{-4}$}} & \makebox{\textbf{1.00 \rpm 0}} & \makebox{6$e^{-4}$ \rpm 4$e^{-4}$} 
\\  
\midrule 
\addlinespace
{DistilBERT} & \makebox{0.91 \rpm 0.01} & \makebox{0.38 \rpm 0.01} & \makebox{0.92 \rpm 0.01} & \makebox{0.39 \rpm 0.01} & \makebox{0.92 \rpm 0.01} & \makebox{0.38 \rpm 0.01} 
\\  
\emph{+Cosreg} & \makebox{0.91 \rpm 0.01} & \makebox{0.39 \rpm 0.01} & \makebox{0.92 \rpm 0.01} & \makebox{0.38 \rpm 0.01} & \makebox{0.92 \rpm 0.01} & \makebox{0.38 \rpm 0.01} 
\\  
\emph{+Spectrum-Pol} & \makebox{\textbf{1.00 \rpm 0.01}} & \makebox{0.012 \rpm 0.016} & \makebox{\textbf{1.00 \rpm 0}} & \makebox{\textbf{7$e^{-4}$ \rpm 5$e^{-4}$}} & \makebox{\textbf{1.00 \rpm 0}} & \makebox{\textbf{11$e^{-4}$ \rpm 9$e^{-4}$}} 
\\  
\emph{+Spectrum-Exp} & \makebox{\textbf{1.00 \rpm 0.01}} & \makebox{\textbf{0.009 \rpm 0.010}} & \makebox{\textbf{1.00 \rpm 0}} & \makebox{\textbf{7$e^{-4}$ \rpm 5$e^{-4}$}} & \makebox{\textbf{1.00 \rpm 0}} & \makebox{\textbf{11$e^{-4}$ \rpm 9$e^{-4}$}} 
\\  
\midrule
  & \multicolumn{2}{c}{RTE} & \multicolumn{2}{c}{WNLI} & \multicolumn{2}{c}{STS-B} 
 \\
\cmidrule(lr){2-3} \cmidrule(lr){4-5} \cmidrule(lr){6-7} 
{Model} & {$I_1 (\uparrow)$} & {$I_2 (\downarrow)$} & {$I_1 (\uparrow)$} & {$I_2 (\downarrow)$} & {$I_1 (\uparrow)$} & {$I_2 (\downarrow)$} 
\\
\midrule  
{BERT} & \makebox{0.92 \rpm 0.01} & \makebox{0.39 \rpm 0.02} & \makebox{0.91 \rpm 0.01} & \makebox{0.39 \rpm 0.02} & \makebox{0.95 \rpm 0} & \makebox{0.22 \rpm 0.01} 
\\  
\emph{+Cosreg} & \makebox{0.92 \rpm 0.01} & \makebox{0.40 \rpm 0.03} & \makebox{0.91 \rpm 0.01} & \makebox{0.40 \rpm 0.01} & \makebox{0.95 \rpm 0.01} & \makebox{0.23 \rpm 0.01} 
\\  
\emph{+Spectrum-Pol} & \makebox{\textbf{1.00 \rpm 0}} & \makebox{\textbf{2$e^{-4}$ \rpm 1$e^{-4}$}} & \makebox{\textbf{1.00 \rpm 0}} & \makebox{\textbf{1$e^{-4}$ \rpm 2$e^{-4}$}} & \makebox{\textbf{1.00 \rpm 0}} & \makebox{0.002 \rpm 0} 
\\
\emph{+Spectrum-Exp} & \makebox{\textbf{1.00 \rpm 0}} & \makebox{3$e^{-4}$ \rpm 2$e^{-4}$} & \makebox{\textbf{1.00 \rpm 0}} & \makebox{2$e^{-4}$ \rpm 3$e^{-4}$} & \makebox{\textbf{1.00 \rpm 0}} & \makebox{\textbf{13$e^{-4}$ \rpm 6$e^{-4}$}} 
\\
\emph{+Flow} & \makebox{0.92 \rpm 0.01} & \makebox{0.39 \rpm 0.01} & \makebox{0.92 \rpm 0.01} & \makebox{0.39 \rpm 0.02} & \makebox{0.95 \rpm 0.01} & \makebox{0.23 \rpm 0.01} 
\\  
\midrule 
\addlinespace
{RoBERTa} & \makebox{0.91 \rpm 0.01} & \makebox{0.40 \rpm 0.01} & \makebox{0.91 \rpm 0.01} & \makebox{0.39 \rpm 0.01} & \makebox{0.95 \rpm 0.01} & \makebox{0.23 \rpm 0.01} 
\\  
\emph{+Cosreg} & \makebox{0.91 \rpm 0} & \makebox{0.41 \rpm 0} & \makebox{0.91 \rpm 0.01} & \makebox{0.40 \rpm 0.01} & \makebox{0.95 \rpm 0} & \makebox{0.23 \rpm 0.01} 
\\  
\emph{+Spectrum-Pol} & \makebox{\textbf{1.00 \rpm 0}} & \makebox{\textbf{3$e^{-4}$ \rpm 2$e^{-4}$}} & \makebox{\textbf{1.00 \rpm 0}} & \makebox{\textbf{3$e^{-4}$ \rpm 1$e^{-4}$}} & \makebox{\textbf{1.00 \rpm 0}} & \makebox{\textbf{7$e^{-4}$ \rpm 3$e^{-4}$}} 
\\  
\emph{+Spectrum-Exp} & \makebox{\textbf{1.00 \rpm 0}} & \makebox{\textbf{3$e^{-4}$ \rpm 2$e^{-4}$}} & \makebox{\textbf{1.00 \rpm 0}} & \makebox{\textbf{3$e^{-4}$ \rpm 1$e^{-4}$}} & \makebox{\textbf{1.00 \rpm 0}} & \makebox{15$e^{-4}$ \rpm 13$e^{-4}$} 
\\  
\midrule 
\addlinespace
{DistilBERT} & \makebox{0.92 \rpm 0.01} & \makebox{0.38 \rpm 0.01} & \makebox{0.92 \rpm 0} & \makebox{0.39 \rpm 0.01} & \makebox{0.95 \rpm 0} & \makebox{0.22 \rpm 0.01} 
\\  
\emph{+Cosreg} & \makebox{0.92 \rpm 0} & \makebox{0.38 \rpm 0.01} & \makebox{0.92 \rpm 0.01} & \makebox{0.38 \rpm 0.01} & \makebox{0.95 \rpm 0} & \makebox{0.22 \rpm 0.01} 
\\  
\emph{+Spectrum-Pol} & \makebox{\textbf{1.00 \rpm 0}} & \makebox{\textbf{2$e^{-4}$ \rpm 3$e^{-4}$}} & \makebox{\textbf{1.00 \rpm 0}} & \makebox{\textbf{1$e^{-4}$ \rpm 2$e^{-4}$}} & \makebox{\textbf{1.00 \rpm 0}} & \makebox{\textbf{9$e^{-4}$ \rpm 1$e^{-4}$}}
\\  
\emph{+Spectrum-Exp} & \makebox{\textbf{1.00 \rpm 0}} & \makebox{\textbf{2$e^{-4}$ \rpm 3$e^{-4}$}} & \makebox{\textbf{1.00 \rpm 0}} & \makebox{\textbf{1$e^{-4}$ \rpm 2$e^{-4}$}} & \makebox{\textbf{1.00 \rpm 0}} & \makebox{\textbf{9$e^{-4}$ \rpm 1$e^{-4}$}} 
\\  
\midrule
 & \multicolumn{2}{c}{QNLI} & \multicolumn{2}{c}{MNLI} & \multicolumn{2}{c}{QQP} \\
\cmidrule(lr){2-3} \cmidrule(lr){4-5} \cmidrule(lr){6-7} 
{Model} & {$I_1 (\uparrow)$} & {$I_2 (\downarrow)$} & {$I_1 (\uparrow)$} & {$I_2 (\downarrow)$} & {$I_1 (\uparrow)$} & {$I_2 (\downarrow)$} \\
\midrule  
{BERT} & \makebox{0.92 \rpm 0.01} & \makebox{0.39 \rpm 0.01} & \makebox{0.93 \rpm 0.01} & \makebox{0.32 \rpm 0} & \makebox{0.92 \rpm 0.01} & \makebox{0.39 \rpm 0.01} \\  
\emph{+Cosreg} & \makebox{0.92 \rpm 0.01} & \makebox{0.39 \rpm 0.01} & \makebox{0.93 \rpm 0.01} & \makebox{0.32 \rpm 0.01} & \makebox{0.9 \rpm 0} & \makebox{0.39 \rpm 0.01}
\\  
\emph{+Spectrum-Pol} & \makebox{0.99 \rpm 0.01} & \makebox{0.06 \rpm 0.02} & \makebox{0.95 \rpm 0.01} & \makebox{0.21 \rpm 0.04} & \makebox{0.92 \rpm 0.02} & \makebox{0.39 \rpm 0.06}
\\
\emph{+Spectrum-Exp} & \makebox{\textbf{1.00 \rpm 0}} & \makebox{\textbf{5$e^{-4}$ \rpm 1$e^{-4}$}} & \makebox{\textbf{0.98 \rpm 0.01}} & \makebox{\textbf{0.08 \rpm 0.03}} & \makebox{\textbf{0.97 \rpm 0.03}} & \makebox{\textbf{0.12 \rpm 0.12}}
\\
\emph{+Flow} & \makebox{0.92 \rpm 0.01} & \makebox{0.39 \rpm 0.01} & \makebox{0.93 \rpm 0} & \makebox{0.31 \rpm 0} & \makebox{0.92 \rpm 0.01} & \makebox{0.39 \rpm 0.01}
\\
\midrule 
\addlinespace
{RoBERTa} & \makebox{0.91 \rpm 0.01} & \makebox{0.40 \rpm 0.01} & \makebox{0.93 \rpm 0.01} & \makebox{0.32 \rpm 0} & \makebox{0.92 \rpm 0.01} & \makebox{0.39 \rpm 0}
\\  
\emph{+Cosreg} & \makebox{0.92 \rpm 0.01} & \makebox{0.40 \rpm 0.01} & \makebox{0.93 \rpm 0.01} & \makebox{0.93 \rpm 0.01} & \makebox{0.32 \rpm 0.01} & \makebox{0.39 \rpm 0}
\\  
\emph{+Spectrum-Pol} & \makebox{\textbf{1.00 \rpm 0}} & \makebox{\textbf{0.005 \rpm 0.003}} & \makebox{0.96 \rpm 0.03} & \makebox{0.15 \rpm 0.13} & \makebox{\textbf{0.99 \rpm 0.2}} & \makebox{\textbf{0.04 \rpm 0.07}}
\\  
\emph{+Spectrum-Exp} & \makebox{\textbf{1.0 \rpm 0.01}} & \makebox{0.012 \rpm 0.015} & \makebox{\textbf{0.98 \rpm 0.01}} & \makebox{\textbf{0.10 \rpm 0.04}} & \makebox{\textbf{0.99 \rpm 0.01}} & \makebox{\textbf{0.04 \rpm 0.06}}
\\  
\midrule 
\addlinespace
{DistilBERT} & \makebox{0.92 \rpm 0} & \makebox{0.38 \rpm 0.01} & \makebox{0.93 \rpm 0.01} & \makebox{0.32 \rpm 0} & \makebox{0.92 \rpm 0.1} & \makebox{0.38 \rpm 0.01} \\  
\emph{+Cosreg} & \makebox{0.92 \rpm 0.01} & \makebox{0.39 \rpm 0.01} & \makebox{0.93 \rpm 0} & \makebox{0.32 \rpm 0} & \makebox{0.992 \rpm 0.01} & \makebox{0.39 \rpm 0.01} \\  
\emph{+Spectrum-Pol}& \makebox{0.99 \rpm 0.01} & \makebox{0.03 \rpm 0.04} & \makebox{0.93 \rpm 0.01} & \makebox{0.29 \rpm 0.01} & \makebox{0.93 \rpm 0.03} & \makebox{0.36 \rpm 0.17} \\  
\emph{+Spectrum-Exp} & \makebox{\textbf{1.00 \rpm 0.01}} & \makebox{\textbf{0.02 \rpm 0.03}} & \makebox{\textbf{0.97 \rpm 0.1}} & \makebox{\textbf{0.13 \rpm 0.01}} & \makebox{\textbf{0.95 \rpm 0.01}} & \makebox{\textbf{0.25 \rpm 0.01}} \\
\bottomrule 
\end{tabular}}
\caption{Isotropy of the embedding space of the different transformer model and calibration method combinations on GLUE tasks.}  
\label{tab:glue_isotropy}
\end{table*} 

The isotropy metrics $I_1$ and $I_2$ show the opposite trend to the performance metrics. An improvement in isotropy reflects a decrease in downstream performance. This way, we see that across models and tasks, cosine regularization and flow calibration (for BERT) have a small impact on isotropy and that the performance of the models calibrated with these techniques is close to the that of the uncalibrated models. On the other hand, spectrum control produces a very significant increase in isotropy, with many tasks reaching a $I_1$ of $1.00$; while in Table~\ref{tab:glue} we see how it produces strong performance degradation. This, further suggests a negative relation between isotropy and the downstream performance of transformers.

\clearpage
\section{Complete Summarization Results}\label{app:sum}

Here we report the complete summarization results, including the ROUGE-2 and ROUGE-L metrics, omitted in the main text.

\begin{table}[ht]
\centering
\resizebox{\textwidth}{!}{\begin{tabular}{@{}l*{7}{S[]}@{}}
\toprule
 & \multicolumn{6}{c}{CNN / Daily Mail}\\
\cmidrule(lr){2-7}
{Model} & {R-1 $(\uparrow)$} & {R-2 $(\uparrow)$} & {R-L $(\uparrow)$} & {$I_2 (\uparrow)$} & {$I_2 (\downarrow)$} & {Time (min)}\\
\midrule  
{BART} & \makebox{\textbf{38.21 \rpm 0.05}} & \makebox{\textbf{17.62 \rpm 0.03}} & \makebox{\textbf{27.06 \rpm 0.08}} & \makebox{0.95 \rpm 0.01} & \makebox{0.25 \rpm 0} & \makebox{246 \rpm 8}\\  

\emph{+Cosreg} & \makebox{\textbf{38.21 \rpm 0.05}} & \makebox{\textbf{17.62 \rpm 0.03}} & \makebox{\textbf{27.06 \rpm 0.08}} & \makebox{0.95 \rpm 0.01} & \makebox{0.25 \rpm 0} & \makebox{240 \rpm 8}\\ 

\emph{+Spectrum-Pol} & \makebox{37.36 \rpm 0.08} & \makebox{16.60 \rpm 0.08} & \makebox{25.26 \rpm 0.09} & \makebox{\textbf{0.99 \rpm 0}} & \makebox{\textbf{0.04 \rpm 0}} & \makebox{245 \rpm 20}\\   

\emph{+Spectrum-Exp} & \makebox{37.43 \rpm 0.08} & \makebox{16.62 \rpm 0.01} & \makebox{26.30 \rpm 0.05} & \makebox{\textbf{0.99 \rpm 0}} & \makebox{\textbf{0.04 \rpm 0}} & \makebox{230 \rpm 18}\\  

\midrule
{BART (small dataset)} & \makebox{\textbf{36.56 \rpm 0.25}} & \makebox{\textbf{15.62 \rpm 0.07}} & \makebox{\textbf{25.05 \rpm 0.07}} & \makebox{\textbf{0.94 \rpm 0}} & \makebox{\textbf{0.25 \rpm 0}} & \makebox{17 \rpm 0} \\  
\emph{+Flow} & \makebox{36.15 \rpm 0.30} & \makebox{15.40 \rpm 0.23} & \makebox{24.79 \rpm 0.19} & \makebox{\textbf{0.94 \rpm 0}} & \makebox{\textbf{0.25 \rpm 0}} & \makebox{95 \rpm 2}\\ 
\bottomrule
\end{tabular}}
\caption{Complete BART summariation performance, embedding space isotropy and fine-tuning time per epoch using different calibration methods on the CNN / DailyMail dataset.
Due to computational cost, the flow calibration method was tested on a smaller version of the dataset with $20\,000$ articles.}  
\label{tab:sum_complete}
\end{table}  

The performance in terms of ROUGE-2 and ROUGE-L scores follows the same patterns as ROUGE-1. Similar to language understanding and machine translation, increasing isotropy does not improve performance. 

\end{document}